# The Potential and Pitfalls of using a Large Language Model such as ChatGPT or GPT-4 as a Clinical Assistant

Jingqing Zhang[1], Kai Sun[1,2], Akshay Jagadeesh[1], Mahta Ghahfarokhi[1], Deepa Gupta[1], Ashok Gupta[1], Vibhor Gupta[1], Yike Guo[1,2,3] *

[1] Pangaea Data Limited, UK, USA
[2] Data Science Institute, Imperial College London, London, UK
[3] Hong Kong University of Science and Technology, Hong Kong SAR, China
* Correspondence to yguo@pangaeadata.ai

## Abstract

Recent studies have demonstrated promising performance of ChatGPT and GPT-4 on several medical domain tasks. However, none have assessed its performance using a large-scale real-world electronic health record database, nor have evaluated its utility in providing clinical diagnostic assistance for patients across a full range of disease presentation. We performed two analyses using ChatGPT and GPT-4, one to identify patients with specific medical diagnoses using a real-world large electronic health record database and the other, in providing diagnostic assistance to healthcare workers in the prospective evaluation of hypothetical patients. Our results show that GPT-4 across disease classification tasks with chain of thought and few-shot prompting can achieve performance as high as 96% F1 scores. For patient assessment, GPT-4 can accurately diagnose three out of four times. However, there were mentions of factually incorrect statements, overlooking crucial medical findings, recommendations for unnecessary investigations and overtreatment. These issues coupled with privacy concerns, make these models currently inadequate for real world clinical use. However, limited data and time needed for prompt engineering in comparison to configuration of conventional machine learning workflows highlight their potential for scalability across healthcare applications.

## Introduction

Large language models (LLMs) play a fundamental role in natural language processing (NLP) and have revolutionised the development of artificial intelligence (AI) systems, as well as the dynamics between AI and human interaction[1]. BERT (Bidirectional Encoder Representations from Transformers)[2], PEGASUS[3], T5 (Text-to-Text Transfer Transformer)[4], and the GPT

(Generative Pre-trained Transformer) models[5] have significantly advanced the field. ChatGPT (GPT-3.5-turbo) is a state-of-the-art LLM developed by OpenAI, surpassing previous models in various benchmarks excelling in tasks such as text generation, language translation, and summarization[6].

In the medical domain, recent publications have demonstrated promising performance of ChatGPT on several tasks, including consultation[7,8], support in clinical workflows such as providing decision support[9], generation of patient notes[10] and discharge summaries[11], simplification of patient radiology reports,[12] medical education applications,[13,14] and several other applications[15]. However, the majority of these studies only provided generic comments about the potential applications of ChatGPT, with only a handful of exemplar use-cases of ChatGPT in specific clinical scenarios, which don't provide a clear picture of ChatGPT's capability[15]. Few studies have provided thorough qualitative and quantitative evaluation of ChatGPT's responses to United States Medical Licensing Exam (USMLE) questions [16,17], or other specialist physician generated ones[18]. However, none have assessed its performance on identification of large target patient cohorts using a large-scale real-world electronic health record database, nor have evaluated its utility in providing clinical assistance for patient evaluation across a full range of a disease (from pre-disease states, to typical and atypical presentations of frank disease, and disease complications). GPT-4, a recent (released March 2023) successor of ChatGPT, exhibits human-level performance on various professional and academic benchmarks[19], and surpasses its predecessor in its performance in a suite of medical benchmark datasets[20].

This study aims at evaluating the strengths and limitations of using ChatGPT and GPT-4, as potential clinical assistants to aid human healthcare professionals. We achieved this with two related but distinct objectives. First, we explore their potential utility in the identification of a large number of suitable patients either for targeted inclusion into prospective research studies, or for the analysis of their pre-existing electronic health record (EHR) data. We do so by evaluating ChatGPT/GPT-4 model performance in classification of patients diagnosed with specific medical conditions including three highly prevalent diseases, namely, Chronic Obstructive Pulmonary Disease (COPD), Chronic Kidney Disease (CKD), and infections by Herpes Simplex Virus (HSV), one rare disease, Primary Biliary Cirrhosis (PBC), and one relatively hard to diagnose disease, cancer cachexia. For this task, we employed a subset of (physician) gold labelled data from the publicly available Medical Information Mart for Intensive Care (MIMIC-III) dataset[21]. To the best of the author's knowledge this is the first study to assess the performance of ChatGPT/GPT-4 in the identification of target patients from unstructured EHR data. Second, using COPD as a case study, we systematically evaluate model utility in providing clinical assistance to patient evaluation. We created a dataset of 31 COPD (and other closely related) case scenarios, sourcing them from previously published case reports[22], British Medical Journal (BMJ) exemplar cases[23], and fictitious cases based on information provided in the Global Initiative for Chronic Obstructive Lung Disease (GOLD) guidelines[24]. We then defined an eight-step clinical care pathway relevant to a physician general practice setting and framed 400 unique questions based on the curated case scenarios that would aid in patient assessment at each step of the pathway. Model responses to these questions were evaluated by licensed physicians using a scoring schema. Our results showed that the performance of GPT-4 across disease classification tasks was good (≥ 75% F1 scores), consistently matching or out-performing the baseline two step 'feature extraction + prediction/rules' models. For patient assessment, GPT-4 was accurate in arriving at a diagnosis three out of four times.

# Results

## Comparison of model performance in binary disease classification

For a sample of EHRs (admissions) from MIMIC-III, we created training and testing sets with physician assigned gold labels. For each disease, gold sets included 200 EHRs (380 for cancer cachexia), with similar proportions of positive and control cases across the gold training and testing sets (Table 1, Methods).

Overall, the performance of GPT-4 across binary disease classification was ≥ 75% (F1 score), with near similar or slightly better performance compared to the corresponding disease-specific 'Extraction + Rules/Prediction' models (Table 1). GPT-4 performance was the lowest for the HSV use-case with an F1 score of 74.70% (with an absolute difference of 17.25% points lower than corresponding the 'Extraction + Rules' model), and the highest for the PBC use-case (with an absolute difference of 3.75% points higher than the corresponding 'Extraction + Rules' model). The relatively lower performance of the GPT-4 model for HSV classification can be attributed to a lower sensitivity of this model compared to the corresponding 'Extraction + Rules' models. This was because patient records with mentions of HSV disease only in those clinical notes apart from the discharge summary, were 'missed out' by the GPT-4 model, as owing to token limits, we were limited to only discharge summary notes as input for the GPT-4 model. This was in contrast to all the relevant patient notes used as input for the 'Extraction + Rules' models. For four of the five diseases, GPT-4 demonstrated superior performance compared to ChatGPT in terms of F1 scores, exhibiting an absolute increase of up to 10%. For the cancer cachexia use-case, ChatGPT F1 scores appeared to be on par with the 'Extraction + Rules/Prediction' models and higher than GPT-4 (nearly 4% absolute difference). However, upon further examination it appears that the ChatGPT model was more likely to classify patients as positive even when truly negative (predicting over 90% of the testing set as positive), and thus has a near perfect recall (96%); the relatively high precision (72%) appears to be an artefact owing to the similar prevalence of cancer cachexia in our testing set (69%).

**Table 1** Evaluation scores obtained by extraction with rules/prediction models, and GPT-4 on the gold standard testing set across the five diseases of interest.

| Disease | Model | Evaluation Metrics | | |
|---|---|---|---|---|
| | | Precision | Recall | F1 Score |
| **COPD** | Extraction + | 95.45% | 1 | **97.67%** |

|  | Rules |  |  |  |
|---|---|---|---|---|
|  | Extraction + Prediction | 96.77% | 95.24% | 96% |
|  | ChatGPT | 84% | 94% | 89% |
|  | GPT-4 | 98.30% | 93.7% | 96% |
| **CKD** | Extraction + Rules | 98.25% | 82.35% | 89.60% |
|  | Extraction + Prediction | 98.33% | 86.76% | **92.19%** |
|  | ChatGPT | 77% | 83% | 80% |
|  | GPT-4 | 98.18% | 79.41% | 87.80% |
| **PBC** | Extraction + Rules | 79.17% | 86.36% | 82.61% |
|  | Extraction + Prediction | 70% | 95.45% | 80.77% |
|  | ChatGPT | 84.21% | 72.73% | 78.05% |
|  | GPT-4 | 86.36% | 86.36% | **86.36%** |
| **HSV** | Extraction + Rules | 90.91% | 93.02% | **91.95%** |
|  | Extraction + Prediction | 68.42% | 90.70% | 78% |
|  | ChatGPT | 48% | 95% | 64% |
|  | GPT-4 | 79.49% | 70.45% | **74.70%** |
| **Cancer Cachexia** | Extraction + Rules | 98.55% | 41.98% | 58.87% |
|  | Extraction + Prediction | 89.05% | 75.31% | 81.61% |
|  | ChatGPT | 72.43% | 95.68% | **82.45%** |
|  | GPT-4 | 82.88% | 74.79% | 78.57% |

Numbers in bold highlight the best performing model/s for that particular disease; Abbreviations: COPD - Chronic Obstructive Pulmonary Disease, CKD - Chronic Kidney Disease, PBC - Primary Biliary Cirrhosis, HSV - Herpes Simplex Virus infections. For all our disease classification tasks we set the ChatGPT/GPT-4 model temperature setting to 0, which is recommended for nudging deterministic responses from GPT models. All ChatGPT/GPT-4 models used an elaborate clinical guideline in the prompt.

For three diseases with the best performance (COPD, CKD, and PBC) we provide sensitivity statistics for key model settings and disease classification performance (Table 2). When using the same input clinical guideline for the same task, GPT-4 compared to ChatGPT could be 10 - 18% higher (in terms of absolute F1 scores) in classification performance. For a given model, simply improving the level of detail in the guideline using best prompting practices such as breaking down the process of assigning a class into subtasks (i.e., chain-of-thought) and providing examples for each class (few shot) can improve classification performance by up to nearly 30% (absolute % increase in F1). GPT-4 with an elaborate clinical guideline had the best classification performance (clinical guidelines in prompts in supplementary S1). Temperature, another key model setting, did not appear to have an effect on the overall disease classification performance (when ensembling repeat predictions using statistical mode). However, lower temperatures increased the reliability over repeat predictions (reliability statistics in supplementary S2).

**Table 2** GPT model and level of detail in clinical guideline in input prompt and model performance in COPD, CKD, and PBC use-cases.

| Disease | Model | Clinical Guideline | Evaluation Metrics | | |
|---|---|---|---|---|---|
| | | | Precision | Recall | F1 Score |
| COPD | ChatGPT | Baseline | 69% | 97% | 81% |
| | | Elaborate | 84% | 94% | **89%** |
| | GPT-4 | Baseline | 83% | 1 | 91% |
| | | Elaborate | 98.30% | 93.7% | **96%** |
| PBC | ChatGPT | Baseline | 26% | 82% | 39% |
| | GPT-4 | Baseline | 40% | 95% | 57% |
| | | Elaborate | 86.36% | 86.36% | **86.36%** |
| CKD | ChatGPT | Baseline | 73% | 81% | 77% |
| | GPT-4 | Elaborate | 98.18% | 79.41% | **87.80%** |

Numbers in bold highlight the best performing model/configuration for that particular disease; Abbreviation: COPD - Chronic Obstructive Pulmonary Disease, PBC - Primary Biliary Cirrhosis. For all our disease classification tasks we set the ChatGPT/GPT-4 model temperature setting to 0.

## ChatGPT/GPT-4 in the prospective evaluation of patients

We explored the utility of ChatGPT in providing clinical assistance (refer supplementary S3) in every step of the generic eight-step clinical care pathway (Figure 2). Our results show that there is some variability in scores assigned by different human evaluators when using x-point likert scales for assessment. Overall, the performance of ChatGPT in providing clinical assistance across the eight-steps was good but varied greatly between the domains assessed, achieving a perfect score of 3 (mean across evaluators), in 229 (57.25%) questions in scientific correctness, 299 (74.75%) in the comprehension, retrieval, and reasoning domain, 109 (27.25%) in the content domain, and 329 (82.25%) in the bias domain. There was also wide variation in the performance across the different steps assessed, with better performance in the earlier steps (steps 1 through 5) compared to the latter (step 6 onwards). Across all steps ChatGPT responses show several instances of missing out on pertinent medical information, making factually inaccurate statements, and infrequently may fabricate content presenting opinions as facts. Asking ChatGPT for investigation recommendations (step 4) or pharmacological management recommendations (step 7) can include those considered superfluous for the given patient scenario.

For a subset of 25 questions pertaining to step 5, using a binary marking schema, we evaluated whether ChatGPT and GPT-4 could accurately produce all the new diagnoses in a patient, given clinical features. GPT-4 performed much better than ChatGPT, with 19 (76%) scored as correct compared to 13 (52%). GPT-4 responses that were scored as correct, included the full primary diagnosis along with key likely complications and any other likely diagnosis unrelated to the patient's primary complaints (such as BP measurement showing hypertension etc.). Among the wrong responses, three diagnoses produced by GPT-4 were likely to have been wrong because the terminology changes in COPD disease definitions (particularly regarding Pre-COPD and PRIsm) is likely to have happened after GPT-4 training cut-off dates. Among the other three wrong responses by GPT-4, all four responses did correctly mention all the new diagnoses, however, they additionally mentioned incorrect diagnoses as highly likely and worth investigating.

## Interpreting the GPT-4 decision making process

For interpretability of GPT-4 predicted disease class, the input prompt instructed the model to provide a rationale for the assigned prediction (examples in supplementary S4). Qualitatively, it appears that for correct predictions (true positives and true negatives), the rationales provided for assigning a particular disease class were usually scientifically correct, exhaustive in context with minimal bias, demonstrating appropriate task comprehension, correct information retrieval, and good adherence to the provided clinical guideline. Additionally, in a handful of cases the model provided scientifically accurate (non-obvious) insights into disease classification extracted from input patient EHR based on information not explicitly mentioned in the clinical guideline. However, for incorrect predictions, (false positives and false negatives), rationales sometimes demonstrated poor accurate comprehension of clinical guidelines, and showed also evidence of fabrication, commonly referred to as LLM hallucination. E.g., saying that a patient's medical record explicitly has a mention of a specific diagnosis when in fact they do not.

Among the clinical case scenarios for diagnosis, the GPT-4 model demonstrated a strong understanding of the given patient information and generated rationales that were coherent and well-supported. This indicates that GPT-4 was able to provide explanations that were consistent with the medical context and mostly aligned with scientific accuracy, without displaying any apparent biases.

## Discussion

Identifying target patients from unstructured EHR data is a relatively complex task, as unstructured clinical narratives in EHR data are typically several pages long, with a lot of variability in how information is documented and recorded. Here the model needs to be able to discern the nuances in natural language, medical parlance, and needs to interpret context to accurately identify patients with diseases of interest. The commonly used methods for identification of target patients, currently, is using medical coding systems like International Classification of Disease (ICD). However, literature has highlighted several flaws in the real-world usage of ICD for this task[25], and the author's previous work has demonstrated that classical machine learning (ML) workflows employing 'extraction with rules/prediction' approaches have better performance[26]. Thus in this study we compared the performance of ChatGPT/GPT-4 only against classical ML workflows. We found GPT-4 with few shot prompting to be rather impressive, with near similar or better performance compared to the corresponding disease-specific ML models.

In classical ML workflows SHAP values, among others, are often used to quantify the contribution of input features in the decision-making process in classification tasks, in terms of a numerical measure of impact[27]. The presence of features clinically relevant to the characterisation of the particular disease of interest fosters model interpretability, engenders trust in the model's output, and contributes to the explainability of artificial intelligence (AI). For an illustration of the SHAP values of the top features contributing to disease classification refer to the authors' previously published work.[26] In contrast, in the context of LLMs such as ChatGPT/GPT-4 one could merely prompt the model to generate a natural language explanation, which outlines the rationale for predicting a particular class with respect to adherence with the provided input clinical guideline. Our results highlight that ChatGPT/GPT-4 rationales for correct predictions are scientifically accurate, comprehensive, unbiased, and demonstrate an understanding of the task and the input guidelines. However, for incorrect predictions, the rationales are unsatisfactory, showing inconsistencies, incorrect retrieval or fabrication of information, and a lack of accurate understanding of the diseases of interest. Though SHAP values have been shown to have limitations and challenges, in terms of the assumptions of feature independence, that of feature linearity and output, data distribution, and several others[28], they do appear to be a mathematically more objective measure than the natural language rationale provided by LLMs.

In providing assistance in the prospective evaluation of patients, our results indicate that though impressive, ChatGPT/GPT-4 models, in their current state, are inadequate to be deployed in real world clinical assistance. ChatGPT responses show evidence for occurrences of incorrect statements, overlooking crucial medical findings, recommending excessive clinical investigations that result in resource wastage, and potentially harmful consequences of overtreatment. GPT-4 was incorrect for one out of four diagnostic questions. The process of human adjudication, although essential, is labour-intensive and susceptible to errors, variations, and biases. Therefore, in future studies, it is imperative to explore word network analyses and other unbiased quantitative NLP metrics to comprehensively investigate the quality and intricacies of AI-generated free-text outputs.

## Constraints of using ChatGPT/GPT-4 for healthcare applications

For classical ML workflows, one can perform threshold modification to increase the precision at the cost of recall, or vice versa, based on the requirement of the clinical application[29]. For example, public health officials may choose higher recall for "screening test" applications but practising clinicians may require higher precision for "confirmatory test" applications. Such threshold modifications are essential to tailor algorithmic utility to specific clinical settings. This is not feasible when using GPT models as it is not possible to perform such thresholding for LLM predictions. Furthermore, token limits, not applicable for classical ML models but for LLMs, may prove to be prohibitive (and thus restrict the number of patient notes provided as input), albeit this may be subject to change in the future given mass adoption. Using GPT models for healthcare applications raises concerns regarding data privacy, primarily due to the transmission of data to external servers hosted by commercial organisations like OpenAI. Mitigating this data privacy issue entails the deployment and operation of a local LLM within a secure network environment, however, the infrastructure and running costs can prove to be restrictive, and further, there exists uncertainty of its performance in comparison to a widely utilised and extensively trained model like ChatGPT/GPT-4.

# Conclusion

In this study, we sought to explore the utility of ChatGPT and GPT-4 as potential clinical assistants by evaluating its performance across two related but distinct tasks. First, we evaluated their performance in the binary classification task of identification of patients with specific diseases of interest (COPD, CKD, PBC, HSV infections, and PBC) from real world EHR data. Second, using COPD as a case study, we evaluated model responses in providing clinical assistance to healthcare workers in prospective patient evaluation. GPT-4 across disease classification tasks with chain of thought and few-shot prompting achieved as high as 96% F1 scores. For patient assessment, GPT-4 was able to arrive at correct diagnoses three out of four times. In this current state, these models are inadequate to be deployed for real world clinical assistance as there were mentions of factually incorrect statements, overlooking crucial medical findings, recommendations for unnecessary clinical investigations and overtreatment. Though there may be privacy concerns regarding use of commercial LLMs like GPT, the limited data and time needed to configure ideal prompting strategies in comparison to conventional ML workflows highlights their huge potential for scalability in healthcare applications.

## Methods

We evaluated the performance of ChatGPT and GPT-4 against classical ML workflows including feature extraction with rules/prediction models in binary disease classification tasks for 5 diseases of interest: COPD, CKD, PBC, infection with HSV, and cancer cachexia. An overview of the study methodology is provided in Figure 1.

**Figure 1:** Overview of the study methodology for the patient identification tasks using MIMIC-III

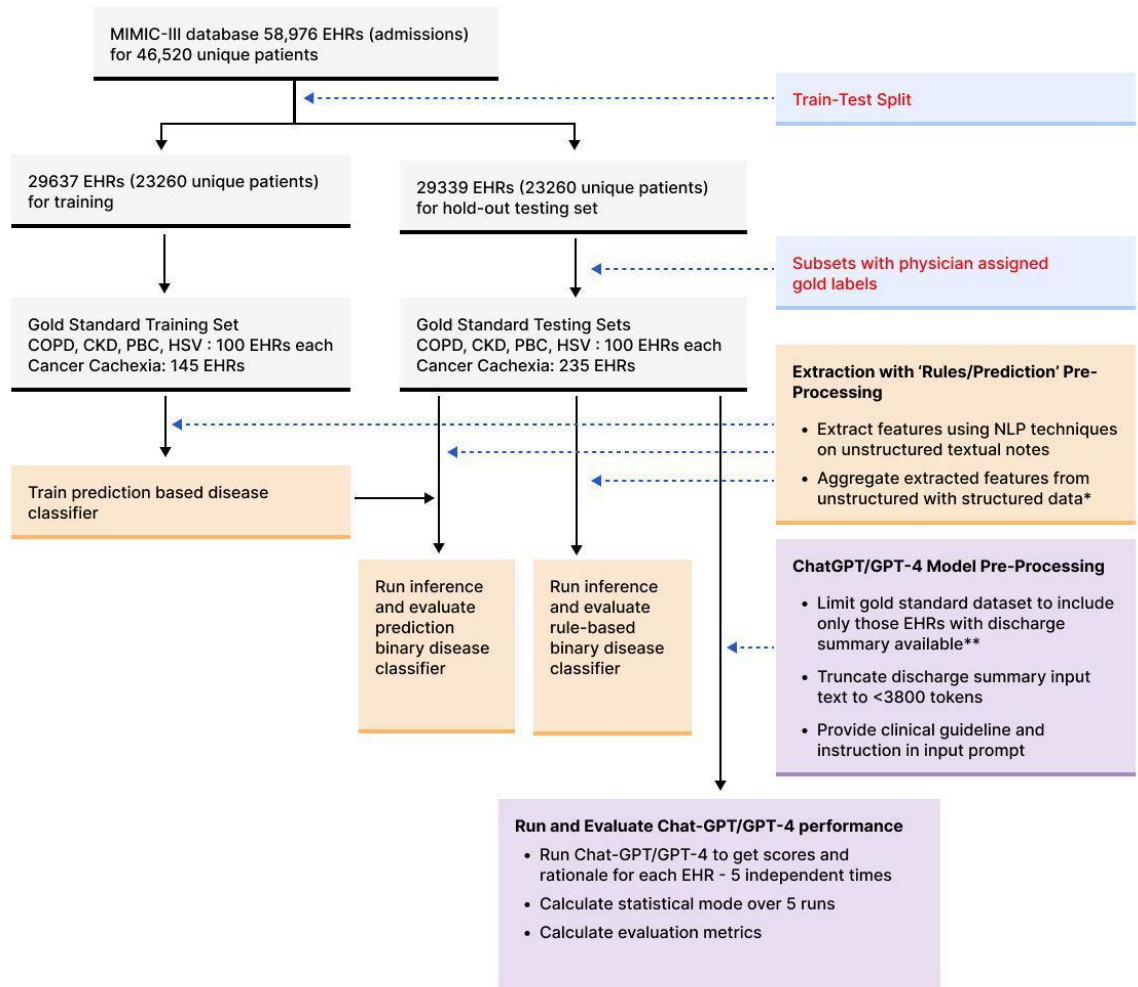

MIMIC-III: Medical Information Mart for Intensive Care, EHRs: Electronic Health Records; * applicable only for the prediction models, and not for rule-based models, ** for the few admissions that lacked a discharge summary, the next most relevant patient note (say physician, nursing, or respiratory notes) were used.

## Dataset: MIMIC-III

In this study we utilised the freely accessible (MIMIC-III) database[21], which collates de-identified, extensive clinical data (both structured as in tabular demographic, chart and lab values, and unstructured as in free text clinical notes like discharge summaries, physician notes, and others) of the 46,520 distinct patients (corresponding to a total 58,976 admissions) admitted to the Intensive Care Units (ICU) of the Beth Israel Deaconess Medical Center, Boston between 2001 and 2012. We performed a random 50% split of all MIMIC-III admissions into training and testing sets ensuring all admissions for a given patient belonged to either training or testing, to avoid data leakage.

To create the gold standard training and testing subsets (Table 3), we followed a sampling strategy to select EHRs for admissions to be manually reviewed and assigned a three-category gold standard label by licensed physicians according to predetermined clinical criteria for each of the 5 diseases of interest. These criteria were developed based on existing medical literature. Initially, we developed an extraction with rules based binary classifiers to identify EHRs containing any mentions of clinical features with positive expression types (affirmed, historical, possible, or advancing) related to each of the 5 diseases of interest, extracted from the clinical notes. This classifier was run on both the training and testing sets labelling each EHR. We then selected EHRs randomly and proportionally from the predicted positive and negative classes from both the training and testing sets. Two clinical experts were tasked with independently evaluating each EHR and assigning a score ranging from 1 to 3, based on clinical guidelines: EHRs without a diagnosis of any of the 5 diseases (Score 1), patients at-risk of the disease or those having pre-disease symptoms (Score 2), and patients with a confirmed diagnosis of the disease (Score 3). Disagreements in scoring were resolved through consensus with a third clinician. The scores were then binarized, with positive labels for EHRs with scores 2 or 3, and control labels for patients with score 1.

**Table 3 Summary statistics of the cases used for disease classification.**

| Disease | Total | Gold Training Set | | Gold Testing Set | |
|---|---|---|---|---|---|
| | | Positive | Control | Positive | Control |
| **COPD** | 200 | 62 | 38 | 63 | 37 |
| **CKD** | 200 | 59 | 41 | 68 | 32 |
| **PBC** | 200 | 24 | 76 | 22 | 78 |
| **HSV** | 200 | 40 | 60 | 43 | 57 |
| **Cancer Cachexia** | 380 | 70 | 75 | 162 | 73 |

Abbreviations: COPD - Chronic Obstructive Pulmonary Disease, CKD - Chronic Kidney Disease, PBC - Primary Biliary Cirrhosis, HSV - Herpes Simplex Virus infections.

## Dataset: Clinical Case Scenarios

We curated clinical vignettes from previously published case reports[22], British Medical Journal (BMJ) COPD exemplar cases[23], and also concocted COPD cases based on information provided in the Global Initiative Chronic Obstructive Lung Disease (GOLD) 2021 guidelines[24]. We chose the 2021 guidelines over the latest 2023 guidelines to accommodate for the fact that the knowledge cut-off date for the ChatGPT model is September 2021. Case scenarios created included typical and atypical presentations of a mix of new and follow-up COPD cases, COPD exacerbation cases, Pre-COPD and Preserved Ratio Impaired Spirometry cases, and other similarly presenting lung conditions like Asthma and lung cancer. All case scenarios were clinically validated by licensed physicians. We developed 31 unique case scenarios of patients with COPD, patients at risk of COPD (Pre-COPD, PRIsm - Preserved Ratio Impaired Spirometry), and other closely related respiratory conditions (Table 4). Further, we constructed and clinically validated a multi-step generic flowchart for assessment of a patient presenting with a set of complaints to a physician general practice (Figure 2). Finally, based on the curated scenarios we built a dataset (provided in supplementary S5) with 400 unique questions. Each question was aimed at patient assessment at a particular step of the clinical care pathway.

**Table 4** Clinical case scenarios used to develop questions for ChatGPT/GPT-4 evaluation

| Type of case scenario | | | Number of case scenarios |
|---|---|---|---|
| **Disease** | **Presentation** | | |
| **COPD** | New diagnosis | Presenting across varying symptom severities across GOLD groups A, B, C, and D with varying disease complications, and comorbidities. | 12 |
| | Previously diagnosed (Follow-up) | Symptomatic (poorly controlled COPD) | 3 |
| | | Asymptomatic (routine follow up) | 3 |
| | Acute exacerbation | Typical exacerbation without presence of infection (with and without peripheral eosinophilia) | 2 |
| | | With bacterial / viral / atypical pneumonia | 3 |
| Pre-COPD | New diagnosis | Presence of COPD symptoms / structural lung lesions with normal FEV/FVC ratios and other spirometry measurements | 1 |
| PRIsm | | Preserved FEV/FVC ratio post-bronchodilation but with other spirometry measures impaired | 2 |
| Asthma | | Late onset with typical asthmatic features | 1 |
| Asthma | | Atypical features (male with significant smoking history, but with positive bronchodilator response with FEV/FVC ratio > 0.7) | 1 |
| COPD-Asthma overlap | | Features of COPD-Asthma overlap as per GOLD guidelines | 2 |
| Lung cancer | | Typical risk factors and presentation | 1 |

Abbreviations: COPD- Chronic obstructive pulmonary disease, PRIsm - Preserved ratio impaired spirometry. GOLD - Global initiative for chronic obstructive lung disease.

**Figure 2** Eight-step flowchart illustrating the evaluation of a patient presenting with a set of complaints in a physician's general practice.

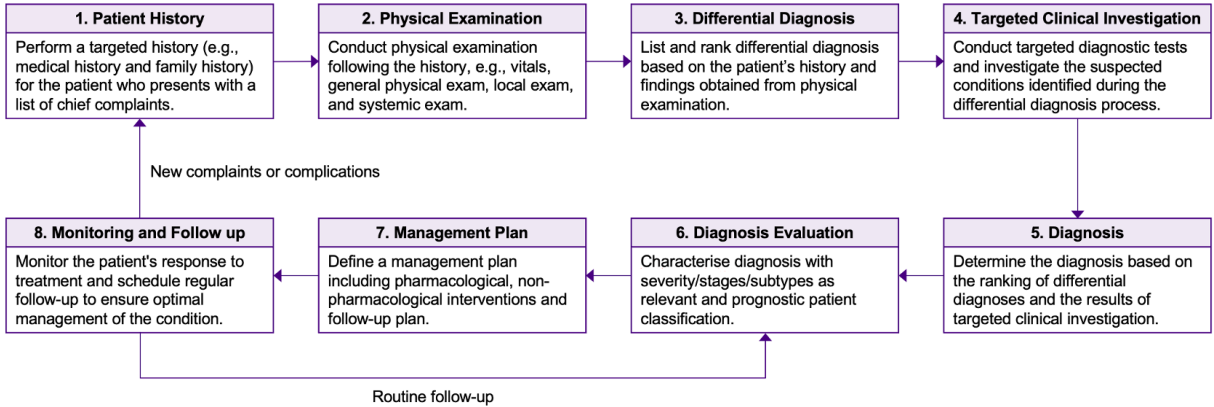

Step 1, includes the necessary inquiries regarding the patient's medical history and specific questions to explore their smoking history in terms of pack-years. Step 2 involves providing anticipated findings on respiratory and cardiovascular system examinations based on the patient's history. In Step 4, targeted clinical investigations are to be recommended based on the patient's medical history and physical examination results. Additionally, if a differential diagnosis or the most likely diagnosis is provided, specific clinical investigations are to be suggested to confirm or rule out each possibility. Clinical interpretation of chest X-ray, chest CT scan, and spirometry tests based solely on investigation reports are also assessed in this step. Step 5 focuses on the diagnosis, where the most likely diagnosis is to be determined based on the patient's history, examination, and investigation reports. A rationale is to be provided, explaining the patient features used to reach the diagnosis. Step 6 entails evaluating the patient following a confirmed diagnosis, which involves determining the COPD GOLD severity stage based on FEV1 values from the spirometry report. Dyspnea grade, COPD Assessment Test (CAT) score, COPD exacerbation risk, and GOLD group (A, B, C, or D) are also to be assessed, with accompanying justifications. Step 7 addresses the management stage, where lifestyle and pharmacological management plans are recommended based on the patient's current medication, vaccination, and other therapies, as well as information obtained from previous steps. Long-term oxygen therapy and adult vaccinations are discussed, with rationales to be provided for each recommendation. Step 8 involves determining the recommended frequency of follow-up based on the patient's diagnosis and medical record. Additionally, if a patient presents with no complaints during follow-up, recommended clinical investigations are to be outlined along with justifications.

**Baseline Classifiers: Extraction with Rules/Prediction**

We used a workflow to identify patients with specific diseases from EHR data as described by Zhang et. al. in 2022[26]. Briefly, it uses unstructured textual data and structured data for each EHR. It leverages a phenotyping NLP algorithm (based on clinical BERT)[30] to capture explicit mentions of clinical concepts (say 'Hypertension'), their synonyms ('high blood Pressure'), abbreviations ('HTN'), numeric values ('BP > 140/90 mm Hg'), and contextual synonyms (for e.g., 'a rise in blood pressure', 'blood pressure above normal range') to annotate clinical concepts in unstructured text by mapping the concept to the Human Phenotype Ontology (HPO). Concepts beyond the scope of the HPO ontology are captured by the Medical Concept Annotation Tool[31], and mapped to the SNOMED-CT (Systematised Nomenclature of Medicine - Clinical Terms) ontology. Additional concepts are identified and annotated using regular expression techniques. The annotated concepts are then assigned an appropriate expression type (for e.g., 'Affirmed', 'Negation', 'Possible') by a context aware classify expression algorithm.

The extracted features (along with the structured data only for the prediction models) form the input to the binary classifier models. The rule-based models use a rule to identify an EHR as positive if they contain any of the relevant extracted clinical concepts for a given disease having a positive expression type. For prediction models, we explored five different machine learning models namely Random Forest, Gradient Boosting, Logistic Regression, Support Vector Machine, and Multi-Layer Perceptron building binary classifiers for the diseases of interest. The best performing model was identified by optimising for the F-1 score, with feature, classifier type selection and hyperparameter optimisation performed using an Automated ML framework[26].

## GPT model configuration

For patient identification task using MIMIC-III, the input prompt to ChatGPT/GPT-4 included the unstructured textual data pertaining to each EHRs' discharge summaries only (truncated to 3800 tokens), along with an abridged disease-specific (baseline) clinical guideline (clinical guidelines in prompts in supplementary S1) and an instruction to assign a 0-1 score along with a rationale for the same. Given the current token limits to ChatGPT/GPT-4 we could not employ any clinical notes apart from discharge summaries. Specific disease use-cases required the assignment of gold standard labels based on relevant clinical notes (discharge summary along with physician, nursing, respiratory, and/or radiology). While we supplied all these pertinent notes as input to (baseline) extraction with rules/prediction models, due to token limitations, we could only provide a discharge summary (or the next most relevant clinical note in few cases where a discharge summary was not available) to ChatGPT/GPT-4. Consequently, the performance of GPT models were inherently limited for EHRs in which crucial information for identifying a patient with a particular disease (say a diagnosis) resided in a note that was not included in the model's input. For patient evaluation across each step of the clinical care pathway, the specific question served as the input prompt to the model.

## Sensitivity Analysis

**Choice of GPT model:** GPT-4 is the latest LLM (released March 14th, 2023) said to supersede its predecessor ChatGPT (GPT-3.5-turbo) in its advanced reasoning capabilities. To assess this, we compare their performance across the disease use-cases, and questions that evaluated the step 5 of the patient clinical care pathway.

**Level of detail in the clinical guideline provided in the input prompt to ChatGPT/GPT-4:** Given that ChatGPT/GPT-4 uses the input prompt as the basis for generating a response, the quality and detail of the prompt can have a significant influence on the response generated by the model. Detailed prompts with specific keywords, facts, or examples are thought to generate more specific and accurate responses while vague or ambiguous prompts may produce poorer responses. To assess the influence of the level of detail in the input prompt on ChatGPT/GPT-4 model responses, using the COPD, CKD, PBC use-cases, we evaluated the model scores using both, a baseline and elaborate clinical guidelines provided in the input prompt. The baseline clinical guideline was a type of zero-shot prompting, as it contained no examples. The elaborate clinical guideline prompt was a type of few-shot prompting, as it contained a few examples (2 to 3) of an excerpt of a hypothetical medical record and expected score, for each of the disease classes. These examples were contrived and not based on any actual medical record in the MIMIC-III dataset. Further, the elaborate guideline was structured into a series of sequential sub-tasks (chain of thought prompting) to be performed.

**Reliability in ChatGPT responses:** In the context of GPT models, temperature is a hyperparameter that controls the randomness or creativity of the generated text. A higher temperature results in more diverse and unpredictable outputs, while a lower temperature leads to more reliable or consistent outputs. To assess the diversity in the score predictions for a given discharge summary and clinical guideline, using the COPD disease use-case, we checked score predictions with a temperature of 0 against 0.2 and 1. To describe the reliability of ChatGPT's score predictions, for each disease use-case, we report the number and proportion of the EHRs that received two or more unique predictions on repeat testing (5 times) given the same clinical guideline and instruction in the prompt, when using the model's default temperature of 1.

## Evaluation

The outputs from extraction with rules/prediction models, and ChatGPT/GPT-4 binary disease class predictions were evaluated against clinician assigned gold standard labels by computing the standard metrics used in classification tasks: precision, recall, and F1 scores.

ChatGPT responses to each question were scored by three licensed physicians independently using an evaluation framework (Table 5). Our framework was a simplified version of that previously published for evaluating LLM generative outputs[32]. For each of the four domains evaluated could be scored on a Likert scale of 1-3. Each response was scored, and then aggregated using mean to produce a final score for a particular domain. Additionally, we also provided a binary correct or wrong based on the answer to the most likely diagnosis question assessed in step 5 of the care pathway. Responses were marked as correct, if the full primary diagnosis (along with relevant complications) and any secondary diagnoses were mentioned.

**Table 5** Criteria used to evaluate ChatGPT responses to the questions asked.

| Domain | | Scoring Criteria |
|---|---|---|
| **Scientific correctness** | 1 | Overall information provided is scientifically incorrect |
| | 2 | Overall information provided is scientifically correct, however, may have minor scientific inaccuracies or information may not be up to date with the latest available society guidelines |
| | 3 | Overall information provided is scientifically correct, without any inaccuracies, up to date with latest available society guidelines |
| **Comprehension** (understanding), **Retrieval** (whether retrieves the required information), and **Reasoning** (whether the provided rationale is appropriate) | 1 | 2 or more of comprehension, retrieval, or reasoning are wrong |
| | 2 | Any 1 of comprehension, retrieval, or reasoning are wrong |
| | 3 | All of comprehension, retrieval, and reasoning are correct |
| **Content** | 1 | Presence of highly inappropriate, harmful, or incorrect content, or Absence of highly relevant content |
| | 2 | No inappropriate, harmful, or incorrect content, however, may be missing some (less important but still) relevant content |
| | 3 | No inappropriate or incorrect content, all relevant content is provided (answer is comprehensive) |
| **Bias** | 1 | Strong evidence of Bias |
| | 2 | May have some evidence of Bias |
| | 3 | No evidence of Bias |

# References

1. Zhao, W. X. *et al.* A Survey of Large Language Models. (2023).
2. Devlin, J., Chang, M.-W., Lee, K. & Toutanova, K. BERT: Pre-training of Deep Bidirectional Transformers for Language Understanding. (2018).
3. Zhang, J., Zhao, Y., Saleh, M. & Liu, P. J. *PEGASUS: Pre-training with Extracted Gap-sentences for Abstractive Summarization*. (2020).
4. Raffel, C. *et al.* Exploring the Limits of Transfer Learning with a Unified Text-to-Text Transformer. (2019).
5. Brown, T. B. *et al.* Language Models are Few-Shot Learners. (2020).
6. Laskar, M. T. R. *et al.* A Systematic Study and Comprehensive Evaluation of ChatGPT on Benchmark Datasets. (2023).
7. Ahn, C. Exploring ChatGPT for information of cardiopulmonary resuscitation. *Resuscitation* **185**, 109729 (2023).
8. Howard, A., Hope, W. & Gerada, A. ChatGPT and antimicrobial advice: the end of the consulting infection doctor? *Lancet Infect Dis* **23**, 405–406 (2023).
9. Rao, A. *et al.* Assessing the Utility of ChatGPT Throughout the Entire Clinical Workflow. *medRxiv* doi:10.1101/2023.02.21.23285886.
10. Ali, S. R., Dobbs, T. D., Hutchings, H. A. & Whitaker, I. S. Using ChatGPT to write patient clinic letters. *Lancet Digit Health* **5**, e179–e181 (2023).
11. Patel, S. B. & Lam, K. ChatGPT: the future of discharge summaries? *Lancet Digit Health* **5**, e107–e108 (2023).
12. Jeblick, K. *et al.* ChatGPT Makes Medicine Easy to Swallow: An Exploratory Case Study on Simplified Radiology Reports. (2022).
13. O'Connor, S. & ChatGPT. Open artificial intelligence platforms in nursing education: Tools for academic progress or abuse? *Nurse Educ Pract* **66**, 103537 (2023).
14. Arif, T. Bin, Munaf, U. & Ul-Haque, I. The future of medical education and research: Is ChatGPT a blessing or blight in disguise? *Med Educ Online* **28**, (2023).
15. Li, J., Dada, A., Kleesiek, J. & Egger, J. ChatGPT in Healthcare: A Taxonomy and Systematic Review. (2023) doi:10.1101/2023.03.30.23287899.
16. Gilson, A. *et al.* How Does ChatGPT Perform on the United States Medical Licensing Examination? The Implications of Large Language Models for Medical Education and Knowledge Assessment. *JMIR Med Educ 2023;9:e45312 https://mededu.jmir.org/2023/1/e45312* **9**, e45312 (2023).
17. Kung, T. H. *et al.* Performance of ChatGPT on USMLE: Potential for AI-assisted medical education using large language models. *PLOS Digital Health* **2**, e0000198 (2023).
18. Johnson, D. *et al.* Assessing the Accuracy and Reliability of AI-Generated Medical Responses: An Evaluation of the Chat-GPT Model. *Res Sq* (2023) doi:10.21203/rs.3.rs-2566942/v1.
19. OpenAI. GPT-4 Technical Report. (2024).
20. Nori, H. *et al. Capabilities of GPT-4 on Medical Challenge Problems*. (2023).
21. Johnson, A. E. W. *et al.* MIMIC-III, a freely accessible critical care database. *Scientific Data 2016 3:1* **3**, 1–9 (2016).

22. Cho, Y. M. *et al.* Work-related COPD after years of occupational exposure. *Ann Occup Environ Med* **27**, (2015).
23. BMJ Best Practice. Chronic obstructive pulmonary disease (COPD) - Case history. (2023).
24. Global Initiative For Chronic Obstructive Lung Disease. *GLOBAL STRATEGY FOR PREVENTION, DIAGNOSIS AND MANAGEMENT OF COPD: 2021 Report*. (2021).
25. Southern, D. A. *et al.* Opportunities and challenges for quality and safety applications in ICD-11: an international survey of users of coded health data. *International Journal for Quality in Health Care* **28**, 129 (2016).
26. Zhang, J. *et al.* A Scalable Workflow to Build Machine Learning Classifiers with Clinician-in-the-Loop to Identify Patients in Specific Diseases. (2022).
27. Lundberg, S. M., Allen, P. G. & Lee, S.-I. A Unified Approach to Interpreting Model Predictions. in *31st Conference on Neural Information Processing Systems (NIPS), Long Beach, CA, USA.* (2017).
28. Kumar, I. E., Venkatasubramanian, S., Scheidegger, C. & Friedler, S. Problems with Shapley-value-based explanations as feature importance measures. in *Proceedings of the 37th International Conference on Machine Learning, Vienna, Austria, PMLR.* (2020).
29. Unal, I. Defining an optimal cut-point value in ROC analysis: An alternative approach. *Comput Math Methods Med* **2017**, (2017).
30. Zhang, J. *et al.* Self-Supervised Detection of Contextual Synonyms in a Multi-Class Setting: Phenotype Annotation Use Case. in *Proceedings of the 2021 Conference on Empirical Methods in Natural Language Processing* 8754–8769 (Association for Computational Linguistics (ACL), 2021). doi:10.18653/V1/2021.EMNLP-MAIN.690.
31. Kraljevic, Z. *et al.* Multi-domain clinical natural language processing with MedCAT: The Medical Concept Annotation Toolkit. *Artif Intell Med* **117**, 102083 (2021).
32. Singhal, K. *et al.* Large Language Models Encode Clinical Knowledge. (2022).